\documentclass[a4paper,10pt,conference]{ieeeconf}
\usepackage{FG2026}

\FGfinalcopy  %

\IEEEoverridecommandlockouts
\usepackage[utf8]{inputenc}
\usepackage[hidelinks]{hyperref}   %
\usepackage{xcolor}
\usepackage{todonotes}             %
\usepackage{lscape}
\usepackage{booktabs}
\usepackage{amssymb}
\usepackage{pifont}
\usepackage{subcaption}
\usepackage{graphicx}
\usepackage{hyperxmp}              %
\usepackage{amsmath}
\usepackage{cite}                  %

\graphicspath{{./figures/}}

\newcommand{\cn}{\textsc{BlEmoRe}}
\newcommand{\accpresence}{\textit{ACC}\textsubscript{presence}}
\newcommand{\accsalience}{\textit{ACC}\textsubscript{salience}}

\def\FGPaperID{****}

\title{\LARGE \bf
Not all Blends are Equal: The \cn{} Dataset of Blended Emotion Expressions with Relative Salience Annotations
}

\begin{document}

\ifFGfinal
\thispagestyle{empty}
\pagestyle{empty}
\author{\parbox{16cm}{\centering
    {\large Tim Lachmann$^1$, Alexandra Israelsson$^1$, 
    Christina Tornberg$^1$, Teimuraz Saghinadze$^2$, Michal Balazia$^3$, Philipp M\"uller$^{4,5}$, Petri Laukka$^6$}\\[2mm]
    {\normalsize
    $^1$ Stockholm University, Sweden, $^2$ Georgian Technical University, Georgia, $^3$ INRIA Université Côte d'Azur, France, $^4$ Max Planck Institute for Intelligent Systems, Germany, $^5$ German Research Center for Artificial Intelligence, Germany, $^6$ Uppsala University, Sweden\\
    tim.lachmann@su.se, 
    petri.laukka@psyk.uu.se
}}
    \thanks{This work was supported by the Marianne and Marcus Wallenberg Foundation (MMW 2018.0059) and the European Union Horizon Europe programme, grant number 101078950.}
}
\else
\author{Anonymous FG2026 submission\\ Paper ID \FGPaperID \\}
\pagestyle{plain}
\fi

\maketitle

\begin{abstract}
Humans often experience not just a single basic emotion at a time, but rather a blend of several emotions with varying salience. Despite the importance of such blended emotions, most video-based emotion recognition approaches are designed to recognize single emotions only. The few approaches that have attempted to recognize blended emotions typically cannot assess the relative salience of the emotions within a blend. This limitation largely stems from the lack of datasets containing a substantial number of blended emotion samples annotated with relative salience. %
To address this shortcoming, we introduce \cn{}, a novel dataset for multimodal (video, audio) \underline{bl}ended \underline{emo}tion \underline{re}cognition that includes information on the relative salience of each emotion within a blend.
\cn{} comprises over 3,000 clips from 58 actors, performing 6 basic emotions and 10 distinct blends, where each blend has 3 different salience configurations (50/50, 70/30, and 30/70). 
Using this dataset, we conduct extensive evaluations of state-of-the-art video classification approaches on two blended emotion prediction tasks: (1) predicting the presence of emotions in a given sample, and (2) predicting the relative salience of emotions in a blend. Our results show that unimodal classifiers achieve up to 29\% presence accuracy and 13\% salience accuracy on the validation set, while multimodal methods yield clear improvements, with ImageBind + WavLM reaching 35\% presence accuracy and HiCMAE 18\% salience accuracy. On the held-out test set, the best models achieve 33\% presence accuracy (VideoMAEv2 + HuBERT) and 18\% salience accuracy (HiCMAE).
In sum, the \cn{} dataset provides a valuable resource to advancing research on emotion recognition systems that account for the complexity and significance of blended emotion expressions.
\end{abstract}

\section{Introduction}
An emotion rarely comes alone.
Instead, humans often experience several emotions at the same time, which is called \textit{blended emotions} (also called compound or mixed emotions) \cite{SpecificityInMixedEmotions, Berrios2015}.
For example, when confronted with a loss that is found to be unjust, we may feel both sadness and anger at the same time.
Blended emotions can also have varying levels of salience, as e.g. a surprise birthday party may lead to a large proportion of happiness but also a smaller proportion of surprise. Self-reports of blended emotions have included both combinations of positive and negative emotions (e.g., happiness and sadness) and emotions that share the same valence (e.g., anger and sadness) \cite{Moeller2018, Oatley1994, Scherer2013}.

The importance of blended emotions in our daily lives is not addressed by the vast majority of video- and audio-based emotion recognition approaches, which usually only focus on recognizing a single emotion at a time \cite{KHARE_REVIEW}.
To the best of our knowledge, dedicated approaches for the recognition of blended emotions only exist for still images~\cite{dong2024bi}.
Video emotion recognition approaches commonly focus on single emotion recognition and, if at all, only evaluate blended emotion recognition as a by-product, without proposing dedicated methods specifically for blended emotion recognition~\cite{li2024pth,sun2024hicmae,zhang2023transformer}.
A major reason for this negligence is the lack of suitable, high-quality datasets. Existing datasets that include blended emotion annotations either have a low number of training samples \cite{muller2015emotion, 10.1145/3293663.3293671}, contain a small number of classes \cite{yang_multimodal_2024, 9110066}, or have a heavily unbalanced class distribution across blended emotions \cite{liu_mafw_2022}. 
In addition, none of the existing blended emotion datasets includes information on the relative salience of the emotions contained in the blend.

We address this limitation by presenting \cn{}, a novel multimodal dataset that provides systematic annotations of the relative salience of emotions in blended expressions. \cn{} contains portrayals of six single emotions (\textit{anger}, \textit{disgust}, \textit{fear}, \textit{happiness}, \textit{sadness}, and \textit{neutral}), along with all pairwise combinations between the non-neutral emotions, resulting in 10 unique combinations. Each blended emotion is further categorized into one of three classes according to the relative salience of the two emotions in the blend: either emotion A is more salient than emotion B, emotion B is more salient than emotion A, or both emotions are equally salient. \cn{} comprises 3,050 clips from 58 actors, with a balanced distribution across single and blended emotion classes, as described below. 

In this paper, we present comprehensive evaluations of state-of-the-art audio- and video encoders with regard to their ability to predict the presence of emotions in a clip, as well as their relative salience.
Our results show that the best performance for emotion presence recognition on the test set is achieved by a multimodal model based on VideoMAEv2 \cite{videomaev2} and HuBERT \cite{hsu2021hubert} embeddings (0.33 accuracy, 16 classes).
The best performance for recognizing the relative salience of emotions contained in a blend was achieved by HiCMAE \cite{sun2024hicmae}, a recent dedicated multi-modal emotion recognition model (0.18 accuracy, 30 classes).
In general, our evaluations show that multimodal approaches outperform their single-modality counterparts.
While performances are clearly above trivial baselines that predict the most likely class, the achieved accuracies indicate a large potential for improvement.
As such, \cn{} is a valuable and challenging resource for the development of automatic blended emotion recognition approaches. The dataset can be accessed on Zenodo\footnote{\url{https://zenodo.org/records/17787362}}.

\section{Related Work}

\subsection{Datasets of Blended Emotions}

\begin{table*}[ht]
    \small
  \caption{Overview over existing publicly available datasets on blended emotion recognition. With \# Samples
we refer to the number of individual video clips, even in the case of frame-wise annotations (such as in MPIIEmo).}
  
  \vspace{-0.25cm}
  \centering
  \begin{tabular}{@{}lrrrrllll@{}}
    \toprule
    Dataset & Participants & \# Samples & \# Single / Blended Samples & Single / Blended Classes & Modalities & Salience\\
    \midrule
    C-EXPR-DB & -    & 400  & -             &  0 + Other    / 12 & Visual, Audio & No\\
    MPIIEmo & 16 & 224 & - &  4 / 6 & Visual, Audio & No \\
    IMED      & 15   & 285                     & 105 / 180     &  6 + Neutral  / 12 & Visual & No\\
    CMED      & -    & 1,050                   & 385 / 665     &  3            /  4 & Visual, ? & No\\
    MD-MER    & 73   & 292  & 219 / 73      &  2 + Baseline /  1 & Visual, ? & No\\
    MAFW      & -    & 8,996                   & 4,938 / 4,058 & 10 + Neutral  / 32 & Visual, Audio & No \\
    \midrule
    \cn{}      & 58   & 3,050                    & 1,390 / 1,660   &  5 + Neutral /  10 & Visual, Audio & Yes \\ 
    \bottomrule
  \end{tabular}
  \label{tab:datasets}
\end{table*}

The first datasets containing expressions of blended emotions were static image datasets~\cite{du2014compound,fabian2016emotionet,li2017reliable,guo2018dominant}.
While CFEE and EmotionNet~\cite{du2014compound,fabian2016emotionet} derived emotion labels from Action Unit detections, RAF-DB and iCV-MEFED~\cite{li2017reliable,guo2018dominant} are fully-human annotated.
RAF-DB employs continuous labelling, meaning each blended emotion is represented as a continuous vector, with its components representing basic emotions. iCV-MEFED considers basic emotions and every pairing of them; hence, every blended emotion consists of dominant and complementary ones. Overall, both datasets recognize potential asymmetry between blends.

More relevant to our work are video datasets of blended emotion expressions.
We present an overview over such datasets in \autoref{tab:datasets}.

\textbf{C-EXPR-DB}~\cite{Kollias_2023_CVPR} consists of 400 videos sourced from YouTube, annotated frame by frame with 12 blended expressions along with other states. The annotations include valence-arousal (VA), action units (AU), speech, facial landmarks, bounding boxes, and facial attributes. Despite the relatively small number of videos, the dataset contains approximately 200,000 frames, equivalent to roughly 13 hours of footage. However, utilizing this data set to explore the interplay between single and blended expressions would be challenging, since it has no video segments with single emotion label. 

\textbf{MPIIEmo}~\cite{muller2015emotion} is a video dataset of acted emotion expressions embedded in short narratives. 
Videos were annotated on a per-frame basis with dimensional and categorical emotion labels.
The total dataset comprises 224 videos, or 252k frames.
Several categorical labels could be given to each video frame, but the authors did not provide an analysis of the relative frequency of blended versus single emotions.

\textbf{IMED}~\cite{10.1145/3293663.3293671} consists of videos featuring 15 Indonesian subjects who were instructed to express 12 blended expressions, 6 basic expressions, and neutral expressions, totaling 285 videos. The recordings were later validated by experts.

\textbf{CMED}~\cite{9110066} is a composite dataset made up of 5 smaller datasets annotated by expert coders, with a total of 1,050 videos. The authors identified 12 blended expressions, 6 basic expressions, and neutral expressions using action unit (AU) coding, as specified by~\cite{du2014compound}. %

\textbf{MD-MER}~\cite{yang_multimodal_2024} is a multimodal data set comprising EEG, GSR, PPG, and frontal face videos from 73 participants. The participants were shown clips selected from a curated list of films from the Stanford film library. The recordings are categorized into three broad emotional states: positive, negative, and mixed, plus baseline. %

\textbf{MAFW}~\cite{liu_mafw_2022} is one of the largest datasets available for blended facial expression recognition. It comprises 10,045 video clips, including audio and verbal scene descriptions, sourced from movies, TV shows, YouTube, and other media. Of these, 8,996 clips are used for the blended expression classification task: 4,938 represent a single expression (across 10 classes + neutral), while 4,058 represent blended expressions (across 32 classes). The data set is highly unbalanced; among the multiple-expression clips, 23 classes represent combinations of two expressions, while 9 classes consist of combinations of three expressions. %

Compared to previous datasets which are either small (IMED, CMED, C-EXPR-DB) or highly unbalanced in the blended emotion class distributions (MAFW), the \cn{} is the second largest dataset of blended emotion expressions and also features a highly balanced class distribution.

\subsection{Methods for Blended Emotion Recognition}

Video emotion recognition is a rapidly growing field.
Multimodal models such as HumanOmni~\cite{zhao2025humanomnilargevisionspeechlanguage}, UMBEnet~\cite{mai2024all}, AVF-MAE++~\cite{wu2025avf}, and VAEmo~\cite{cheng2025vaemo} mostly report better results than video-only models such as FineCliper \cite{chen2024finecliper}, S4D~\cite{chen2025staticdynamicdeeperunderstanding}, and MAE-DFER ~\cite{sun2023mae}.
However, most of these methods were not evaluated on the blended emotion recognition task.

The MAFW-43~\cite{liu_mafw_2022} dataset is the blended emotion recognition dataset that has been most frequently covered in the recent literature. 
However, the models AVF-MAE++~\cite{wu2025avf}, T-MEP~\cite{zhang2023transformer}, HiCMAE~\cite{sun2024hicmae}, and PTH-Net~\cite{li2024pth} fine-tuned on this dataset typically aim for optimal performance across a wide range of datasets, including the MAFW-11 class (non-blended emotion labels). Consequently, they often do not employ specific techniques to induce biases for blended emotions. 
One way to circumvent treating each blended emotion as a separate class, especially for datasets like RAF-DB~\cite{li2017reliable} which have continuous labels, is to use only the base class in the cross-entropy loss. 
More recently, an interesting new objective function, called bi-center loss~\cite{dong2024bi}, has been proposed. This method not only utilizes cross-entropy loss, but also introduces a loss term that anchors the features to their respective centers, corresponding to the base emotions.
Overall, work on dedicated methods for blended emotion recognition remains limited.

\section{The \cn{} Dataset}

\begin{figure*}[h]
    \centering
    \includegraphics[width=0.90\textwidth]{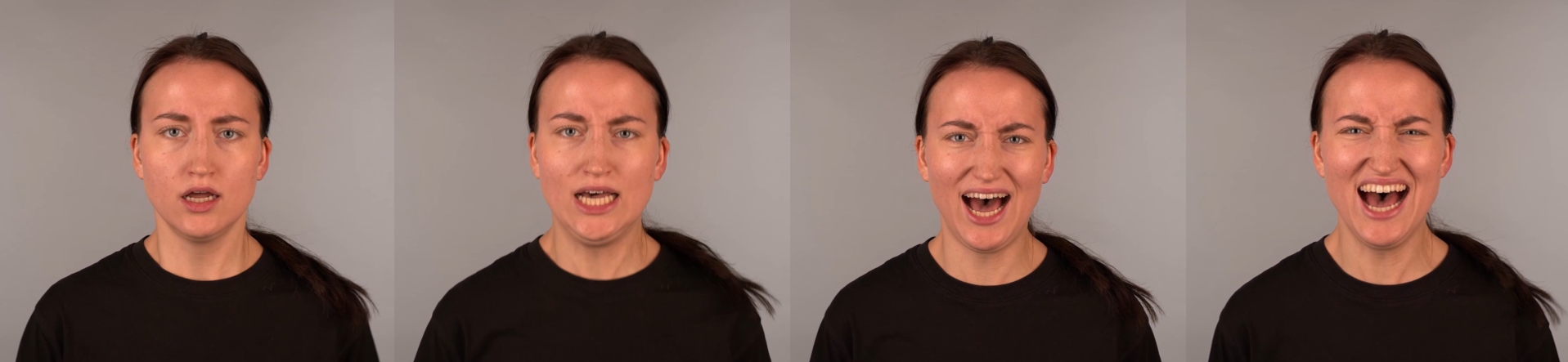}
    \caption{Examples of stills from the video recordings. The actor portrays a combination of anger and fear. Reproduced from~\cite{IsraelssonBlended} under CC BY 4.0.}
    \label{fig:israelsson-figure}
\end{figure*}

We introduce \cn{}, the first publicly available dataset of multimodal blended emotion portrayals with salience annotations. 
\cn{} includes 58 actors who portray a specific range of single and blended emotions through facial expressions, body movements, and vocal sounds. In contrast to most previous efforts, the stimuli were designed to convey blended emotions, rather than being assigned to a blended emotion category based on posthoc annotations. The data recording protocol and initial human emotion recognition experiments on a subset of the recordings were first presented in \cite{IsraelssonBlended}. However, that work did not include any machine learning based emotion recognition experiments. Here, we present the dataset in a format suitable for machine learning applications, along with appropriate evaluation metrics. Ethical approval for the recordings was granted by the Swedish Ethical Review Authority (decision no. 2021-00972).

\subsection{Dataset}
Actors were instructed to express both single emotions (anger, disgust, fear, happiness, sadness, and neutral) and blended emotions consisting of all pairwise combinations of anger, disgust, fear, happiness, and sadness. All pairwise combinations were portrayed in three different blend conditions: 

\begin{itemize}
    \item 50/50 = same amount of both emotions (e.g., 50/50 happiness-sadness: both happiness and sadness are expressed in equal proportions)
    \item 70/30 = the first emotion is more salient than the second emotion (e.g., 70/30 happiness-sadness: conveys mainly happiness blended with a tinge of sadness)
    \item 30/70 = the second emotion is more salient than the first emotion (e.g., 30/70 happiness-sadness: conveys mainly sadness blended with a tinge of happiness)
\end{itemize}

It should be noted that the labels 30/70 and 70/30 do not imply that actors were instructed to express emotions in precise proportional amounts. Instead, these terms indicate that one emotion was intended to be less prominent than the other in the resulting blended expression.

The actors were instructed to convey the emotions as convincingly as possible, as if interacting with another person (the camera), but without using overtly stereotypical expressions. Actors were given definitions and example scenarios for every single emotion \cite{laukkaExpressionRecognitionEmotions2016}, but were free to create their own scenarios for the blended emotions. They were instructed to recall a personally experienced situation that elicited the specified combination and, if possible, to try to re-enter a similar emotional state during the portrayal. 

It was further specified that they should try to express the emotion simultaneously through both the face/body and the voice. For the vocal expressions, they were free to choose any non-linguistic vocalization (e.g., cries, laughter, groans), but no words (including made-up words) were allowed \cite{CrossCulturalLaukka}.  Example stills from a portrayal of a blended emotion are shown in Figure~\ref{fig:israelsson-figure}.

Recordings were conducted in a room with studio lighting and dampened acoustics, using a high-quality camera and microphone. The audio level was calibrated relative to the loudest expected level and then kept constant during the recording session. The camera was placed in front of the actor at a distance of approximately 1.2 m, and the microphone was located 0.5 m above the actor and directed at the actor’s chest (for details, see \cite{IsraelssonBlended}). 

\begin{figure}[h]
    \centering
    \includegraphics[width=0.49\textwidth]{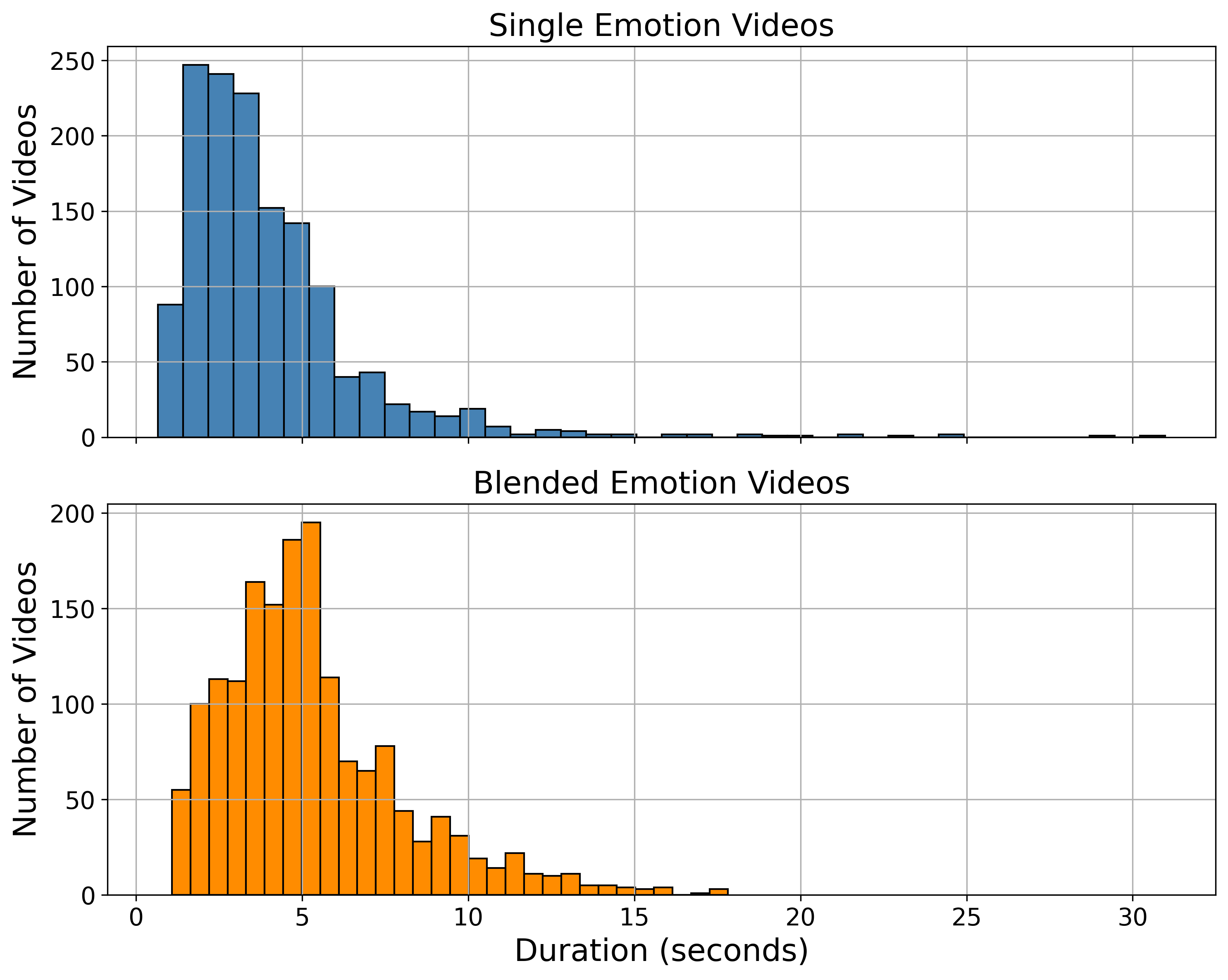}
    \caption{Distribution of video durations in the dataset for single and blended emotions.}
    \label{fig:video-durations}
\end{figure}

\begin{figure}[h]
\centering
\includegraphics[width=0.32\textwidth]{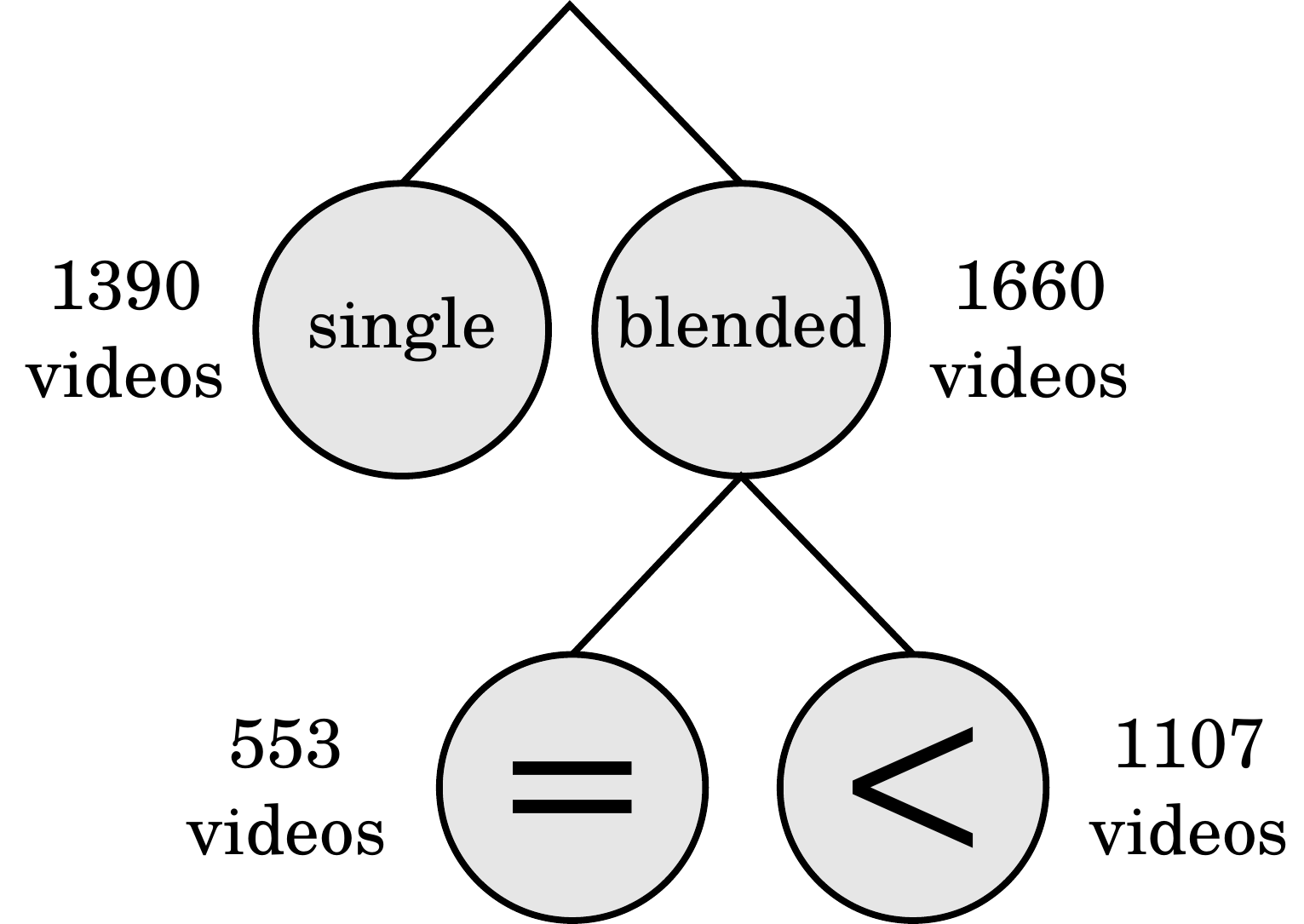}
\caption{Structure of the \cn{} full dataset (train and test partition), which contains single emotions and blended emotions expressed with equal ($=$) and unequal ($<$) salience.} 
\label{fig:BleMoRe-structure}
\end{figure}

We have selected recordings from 58 actors (Male = 28, Female = 30, $M_{age}$ = 36 years, Range = 21 – 77 years), for a total of 1390 recordings of single emotions, and 1660 recordings of blended emotions (see \textbf{Fig~\ref{fig:BleMoRe-structure}}). The duration of the recordings ranges from 1-30 seconds (see \textbf{Fig~\ref{fig:video-durations}}). The dataset labels are based on actor instructions, \cite{IsraelssonBlended} validated a subset of 18 actors through human perception experiments. Human judges performed well above chance on a presence-based accuracy measure comparable to \accpresence{}, achieving a mean accuracy of 0.43 for correctly identifying both emotions in a blend under multimodal (audio-visual) conditions.

 The data were partitioned into a predefined training and test split with no actor overlap: 43 actors are included in the training set and 15 actors in the test set. The test set contains a fixed subset of actors selected to ensure balanced gender representation.

To enable consistent validation, we also provide five predefined cross-validation folds within the training set. Each fold contains a disjoint subset of actors and is approximately balanced in terms of gender and number of samples. This setup ensures that models are evaluated on their ability to generalize to previously unseen individuals.

\subsection{Reference Metrics}
We provide two reference evaluation metrics: \accpresence{} and \accsalience{}.  
\begin{itemize}
    \item $ACC_{presence}$ measures whether the correct label(s) are predicted without errors. A correct prediction must include all present emotions while avoiding false negatives (e.g., predicting only one emotion in a blend) and false positives (e.g., predicting emotions that are not part of the label).
    \item $ACC_{salience}$ extends $ACC_{presence}$ by considering the relative prominence of each emotion. It evaluates whether the predicted proportions accurately reflect the correct ranking, i.e., whether the emotions are equally salient or one is more dominant than the other. This metric only applies to blended emotions.
\end{itemize}
To describe these metrics formally, let's assume both \(k\):th prediction and data-point are represented by six-dimensional vectors such that,
\begin{align*}
p^{(k)}, d^{(k)} \in 
\{\, \gamma e_i + \delta e_j \;|\;
i,j \in \{1,\dots,5\},\, i \neq j,\, \\
(\gamma,\delta)\!\in\!\{(1,0),(0.5,0.5),(0.3,0.7)\}\,\} \cup \{ e_{6} \},
\end{align*}

Where \(e_{1},e_{2}, \dots e_{5}\) are basis vectors corresponding to basic emotions of anger, disgust, fear, happiness, and sadness; \(e_{6}\) is reserved for neutral. The respective coefficients represent relative salience.

We define two auxiliary functions that operationalize the evaluation procedure:
\(\pi : \mathbb{R}^{6} \to \mathbb{R}^{6}\) and \(\sigma : \mathbb{R} \to \mathbb{R}\).
The function \(\pi(\cdot)\) standardizes predictions to disregard salience information, effectively mapping any blended prediction to its presence-only equivalent (e.g., both \(0.3e_i + 0.7e_j\) and \(0.5e_i + 0.5e_j\) become \(0.5e_i + 0.5e_j\)).  
In contrast, \(\sigma(\cdot)\) acts as an indicator function that returns \(1\) if the predicted and target vectors match exactly and \(0\) otherwise.  

\begin{equation*}
\pi(x) =
    \begin{cases}
    e_{i}                 &\text{if } x = e_{i},\\
    0.5e_{i} + 0.5e_{j}   &\text{if } x = \lambda e_{i} + (1-\lambda)e_{j} \text{ and } i \neq j,\\
    0                     &\text{otherwise;}
    \end{cases}
\end{equation*}

\begin{equation*}
\sigma(x) =
    \begin{cases}
    1 &\text{if $x \leq 0$,}\\
    0 &\text{otherwise.}
    \end{cases}
\end{equation*}
Using these definitions, we compute:

\begin{equation*}
ACC_{\text{presence}} = \frac{1}{N} \sum_{k=1}^{N}{\sigma(\| \pi(p^{(k)}) - \pi(d^{(k)}) \|)}
\end{equation*}

\begin{equation*}
ACC_{\text{salience}} = \frac{1}{N} \sum_{k=1}^{N}{\sigma(\| p^{(k)} - d^{(k)} \|)}
\end{equation*}
$ACC_{presence}$ is in effect a generic way to measure (blended) emotion recognition, and is easily comparable to models implemented on other datasets. $ACC_{salience}$ captures the unique feature of our dataset - emotion salience in different blend conditions.

\section{Method}
Our approach extracts features from visual and auditory modalities using pre-trained encoders, followed by classification via feedforward neural networks or a linear layer for multi-label emotion prediction (presence and salience). We detail the encoding methods, feature aggregation, multimodal fusion, label encoding, model configurations, and training pipeline below.

\subsection{Encoding Methods}

We employed eight pre-trained encoders: five for video, two for audio, and one multimodal, spanning frame-level, spatiotemporal, and audio-based representations.

\textbf{Video Encoders.} We used five pre-trained vision models covering both frame-based and spatiotemporal architectures:
\begin{itemize}
    \item \textbf{OpenFace 2.0}~\cite{baltrusaitis2018openface}: Extracts frame-level facial behavior features, including 17 action unit (AU) intensities, 6 head pose parameters, and 8 gaze features (31 features per frame). OpenFace captures fine-grained frame-level facial dynamics.
    \item \textbf{CLIP}~\cite{CLIPradford2021learning}: The vision encoder from CLIP produces frame-level embeddings using contrastive learning between images and text. Each frame is encoded independently.
    \item \textbf{ImageBind}~\cite{girdhar2023imagebind}: Similar to CLIP, ImageBind aligns visual features with multiple sensory modalities. We use its vision encoder to obtain per-frame embeddings.
    \item \textbf{VideoMAE V2}~\cite{videomaev2}: A spatiotemporal encoder operating on 3D video cubes via masked reconstruction. We use the ViT-B/16 variant, which models temporal structure through tube tokens.
    \item \textbf{Video Swin Transformer}~\cite{VideoSwinTransformer}: A hierarchical spatiotemporal model applying shifted 3D window attention to capture both local and global temporal context.
\end{itemize}
\textbf{Audio Encoders.} To extract speech-based representations, we employed two large self-supervised encoders trained on raw 16kHz waveforms:
\begin{itemize}
    \item \textbf{HuBERT}~\cite{hsu2021hubert}: Trained on 60k hours of speech using masked prediction of k-means cluster labels. We use the Large (LL-60k) variant.
    \item \textbf{WavLM}~\cite{WavLM_Chen_2022}: Trained on 94k hours of speech with a combined masked prediction and denoising objective. We use the Large configuration.
\end{itemize}
These audio encoders produce 1024-dimensional frame-level embeddings at 20ms resolution.

\textbf{Multimodal Encoder.} For multimodal representation learning, we directly employed the pre-trained HiCMAE~\cite{sun2024hicmae} model. Following its original setup, we cropped and aligned faces prior to feature extraction and used the resulting fused embeddings for classification without additional architectural modification.

\textbf{Feature Aggregation and Subsampling.} OpenFace, CLIP, ImageBind, HuBERT, and WavLM yield frame-level representations, whereas VideoMAEv2 and Video Swin Transformer produce spatiotemporal representations summarizing multiple frames. We applied two complementary strategies to summarize frame- and spatiotemporal embeddings into fixed-size representations suitable for training:
\begin{itemize}
    \item \textbf{Aggregation:} For frame-level encoders (OpenFace, CLIP, ImageBind, HuBERT, WavLM) and as one variant for spatiotemporal encoders (VideoMAEv2, Video Swin Transformer), we computed seven statistical descriptors per feature dimension: mean, standard deviation, and the 10th, 25th, 50th (median), 75th, and 90th percentiles. These statistics were concatenated into a fixed-size feature vector.
    \item \textbf{Subsampling:} For spatiotemporal encoders (VideoMAEv2, Video Swin Transformer), videos are processed as sequences of fixed-length clips, each consisting of 16 frames. Each clip is passed through the encoder to obtain an embedding summarizing the spatiotemporal information within that segment. We refer to these embeddings as \textit{subsamples}. Under the subsampling strategy, instead of aggregating the subsample embeddings into a single video-level vector, we treat each subsample independently during training. 
\end{itemize}
All features were standardized to a zero mean and unit variance using statistics from the training set. In the case of HiCMAE, we did not make any new inventions in terms of feature representations but stuck to the procedure described in \cite{sun2024hicmae}.

\textbf{Multimodal Fusion.} To assess the benefits of combining modalities, we implemented simple \textbf{early fusion} by concatenating aggregated features from paired video and audio encoders before classification. Rather than exploring all combinations exhaustively, we selected the top-performing visual encoders (VideoMAEv2 and ImageBind) and paired each with both audio encoders (HuBERT and WavLM).

\subsection{Label Encoding}

Each video was annotated with either a single or blended emotion along with relative salience information. 

\begin{itemize}
    \item \textbf{Single-emotion recordings} were encoded as one-hot vectors with a value of 1 at the corresponding emotion index and 0 elsewhere.
    \item \textbf{Blended emotions} were represented as soft probability distributions over two emotions, where the label values correspond to the salience proportions normalized to sum to 1 (e.g., 70\% happiness and 30\% sadness are encoded as [0, 0, 0, 0.7, 0.3, 0]).
\end{itemize}
The resulting target vector for each video has six dimensions, one for each emotion category (anger, disgust, fear, happiness, sadness, and neutral). Each dimension can either have a hard assignment in the case of single-emotion recordings, or soft assignments for blended emotions, except in the case of neutral which never occurs within blends.

\subsection{Classification Models}
For HiCMAE, we adhered to~\cite{sun2024hicmae}, using a single linear layer to project pooled features into emotion probabilities. For other encoders, we trained feedforward neural networks with three configurations:

\begin{itemize}
    \item \textbf{Linear}: A single linear layer mapping input features directly to emotion probabilities.
    \item \textbf{MLP-256}: A network with one hidden layer containing 256 units, followed by ReLU activation.
    \item \textbf{MLP-512}: A network with one hidden layer containing 512 units, followed by ReLU activation.
\end{itemize}
All models produced a six-dimensional output corresponding to the six emotion categories. A softmax activation was applied to the output logits to obtain a probability distribution over the classes.

The models were trained to minimize the Kullback-Leibler (KL) divergence between the predicted softmax distribution and the ground-truth label distribution, reflecting either a hard one-hot vector for single emotions or a soft distribution for blended emotions. For HiCMAE specifically, cross-entropy loss was used as this turned out to outperform KL divergence in validation experiments.

\subsection{Training and Post-processing}

Unimodal and early fusion models (excluding HiCMAE) were trained with the Adam optimizer, with a batch size of 32 for aggregation-based features and 512 for subsampled features. The learning rate was set to $5 \times 10^{-6}$ with weight decay of $1 \times 10^{-3}$. Models were trained with an upper bound of 200 epochs for aggregation-based features and 300 epochs for subsampled features, as preliminary experiments indicated that these settings consistently led to convergence, ensuring that further training would not yield meaningful accuracy improvements despite some instability across epochs. For HiCMAE, the hyperparameters prescribed in \cite{sun2024hicmae} were replicated, and the model was fine-tuned for 50 and 100 epochs using a cosine-learning rate scheduler with a linear warm-up and batch size of 32. Validation accuracy was recorded at each epoch for all setups, and the best model was selected after training was completed. 

The number of training epochs used for the held-out test set was fixed at 100, determined empirically from approximate convergence trends in validation accuracy observed during five-fold cross-validation.

To convert model outputs into discrete emotion presence and salience predictions, we applied a post-processing step involving thresholding. A grid search was conducted on the validation folds to find the optimal thresholds:

\begin{itemize}
    \item \textbf{Presence threshold} ($\alpha$): Determines which emotions are predicted as present.
    \item \textbf{Salience threshold} ($\beta$): Determines whether the blend is classified as equal (50/50) or dominant/subdominant (70/30 or 30/70).
\end{itemize}
During cross-validation $\alpha$ and $\beta$ parameter values were optimized on the validation set. For the held-out test set, $\alpha$ and $\beta$ values were selected from the most successful validation fold and epoch.

In the subsampling approach, predictions were first generated at the subsample level. Final video-level prediction was obtained by averaging the logits across all subsamples before applying the softmax activation and thresholding steps.

\section{Evaluation}

We evaluated the baseline models using five-fold cross-validation on the training set and report performance in terms of presence accuracy (\accpresence{}) and salience accuracy (\accsalience{}). Model selection was based on a composite validation score, computed as the average of presence and salience accuracy. 

For all encoders except HiCMAE, we evaluated three classifier configurations: \textbf{Linear}, \textbf{MLP-256}, and \textbf{MLP-512}. Across these modalities and settings, the \textbf{MLP-512} configuration typically yielded the best validation performance, and results are reported for this configuration. For HiCMAE, we adhered to the authors' prescribed approach using a single linear layer, as detailed in~\cite{sun2024hicmae}.

To contextualize the model performance, we computed trivial baselines based on class distributions. For the single-emotion baseline, always predicting the most frequent single emotion resulted in \accpresence{} of 7.70\% on the training set and 7.40\% on the test set, with \accsalience{} of 0\% by definition. For the blend baseline, always predicting the most frequent blend yielded \accpresence{} of 5.60\% and 5.60\%, and \accsalience{} of 3.50\% and 3.30\% on the training and test sets, respectively.

\subsection{Validation Results}

Table~\ref{tab:val_test_results_full} summarizes the cross-validated performance (mean $\pm$ std) for each encoder, combinations of encoders, and method. Among unimodal encoders, ImageBind yielded the highest composite score (\accpresence{}=0.290, \accsalience{}=0.130), WavLM had the best performance overall among the audio encoders (\accpresence{}=0.265, \accsalience{}=0.121). Multimodal combinations consistently outperformed unimodal variants, with ImageBind + WavLM achieving the best composite score (\accpresence{}=0.345, \accsalience{}=0.170). This suggests that audio and visual modalities carry complementary information that can be leveraged with a simple early fusion strategy. Finally, we evaluated the multimodal HiCMAE model~\cite{sun2024hicmae}. Among all models, HiCMAE achieved the highest salience accuracy (\accsalience{} = 0.180). The best overall HiCMAE performance, reported in Table~\ref{tab:val_test_results_full}, was obtained after 100 epochs using the cross-entropy loss function.

Aggregation-based features generally outperformed subsampled features, see Table~\ref{tab:val_results_subsampling}. For subsampled features, the highest scores were achieved by VideoMAEv2, with \accpresence{}=0.260, \accsalience{}=0.124. This suggests that pooling frame-level or segment-level features into global summary statistics provides more robust representations for downstream prediction.

\begin{table*}[h]
\centering
\caption{Validation (mean $\pm$ std) and test results for aggregation-based features using the MLP-512 model, as well as the HiCMAE model.}
\label{tab:val_test_results_full}
\begin{tabular}{l c c c c}
\toprule
\textbf{Encoder} & \multicolumn{2}{c}{\textbf{Validation}} & \multicolumn{2}{c}{\textbf{Test}} \\
 & \textbf{\accpresence{}} & \textbf{\accsalience{}} & \textbf{\accpresence{}} & \textbf{\accsalience{}} \\
\midrule
CLIP                  & 0.266 $\pm$ 0.021 & 0.105 $\pm$ 0.012 & 0.258 & 0.096 \\
ImageBind             & 0.290 $\pm$ 0.028 & 0.130 $\pm$ 0.008 & 0.261 & 0.087 \\
OpenFace              & 0.228 $\pm$ 0.014 & 0.119 $\pm$ 0.014 & 0.226 & 0.081 \\
VideoMAEv2            & 0.273 $\pm$ 0.025 & 0.106 $\pm$ 0.014 & 0.293 & 0.054 \\
VideoSwin             & 0.225 $\pm$ 0.026 & 0.089 $\pm$ 0.033 & 0.214 & 0.093 \\
HuBERT                & 0.243 $\pm$ 0.023 & 0.104 $\pm$ 0.024 & 0.274 & 0.120 \\
WavLM                 & 0.265 $\pm$ 0.027 & 0.121 $\pm$ 0.012 & 0.311 & 0.084 \\
\midrule
ImageBind + WavLM     & \textbf{0.345} $\pm$ \textbf{0.035} & 0.170 $\pm$ 0.055 & 0.327 & 0.114 \\
ImageBind + HuBERT    & 0.339 $\pm$ 0.023 & 0.158 $\pm$ 0.053 & 0.298 & 0.084 \\
VideoMAEv2 + WavLM    & 0.343 $\pm$ 0.022 & 0.140 $\pm$ 0.028 & 0.332 & 0.102 \\
VideoMAEv2 + HuBERT   & 0.332 $\pm$ 0.016 & 0.138 $\pm$ 0.012 & \textbf{0.332} & 0.114 \\
\midrule
HiCMAE & 0.298 $\pm$ 0.025 & \textbf{0.180} $\pm$ \textbf{0.036} & 0.268 & \textbf{0.180} \\
\midrule
trivial baseline (single emotion) & 0.077 $\pm$ 0.005 & 0.000 $\pm$ 0.000 & 0.074 & 0.000 \\
trivial baseline (blend)          & 0.056 $\pm$ 0.005 & 0.035 $\pm$ 0.003 & 0.056 & 0.033 \\
\bottomrule
\end{tabular}
\end{table*}

\begin{table*}[h]
\centering
\caption{Validation results (mean $\pm$ std) comparing aggregation and subsampling for VideoMAEv2 and VideoSwin.}
\label{tab:val_results_subsampling}
\begin{tabular}{l c c c c}
\toprule
\textbf{Encoder} & \multicolumn{2}{c}{\textbf{Aggregation}} & \multicolumn{2}{c}{\textbf{Subsampling}} \\
 & \textbf{\accpresence{}} & \textbf{\accsalience{}} & \textbf{\accpresence{}} & \textbf{\accsalience{}} \\
\midrule
VideoMAEv2  & \textbf{0.273} $\pm$ \textbf{0.025} & 0.106 $\pm$ 0.014 & 0.260 $\pm$ 0.030 & \textbf{0.124} $\pm$ \textbf{0.027} \\
VideoSwin   & 0.225 $\pm$ 0.026 & 0.089 $\pm$ 0.033 & 0.210 $\pm$ 0.024 & 0.103 $\pm$ 0.020 \\
\bottomrule
\end{tabular}
\end{table*}

\subsection{Test Set Results}

Table~\ref{tab:val_test_results_full} summarizes test performance for the best models. 
The top unimodal model was WavLM (\accpresence{} = 0.311, \accsalience{} = 0.084). 
Among visual models, VideoMAEv2 reached the highest \accpresence{} = 0.293, but lower \accsalience{} = 0.054. Multimodal combinations showed clear benefits: VideoMAEv2 + HuBERT achieved the highest \accpresence{} = 0.332, while HiCMAE obtained the best \accsalience{} =  0.180. There is a general drop in model performance on the test set, even though these models are trained on a larger amount of data (the entire training set). Furthermore, the ranking of different models is not always congruent between validation and test evaluation.

\subsection{Confusion Matrix}

\begin{figure}[h]
    \centering
    \includegraphics[width=0.49\textwidth]{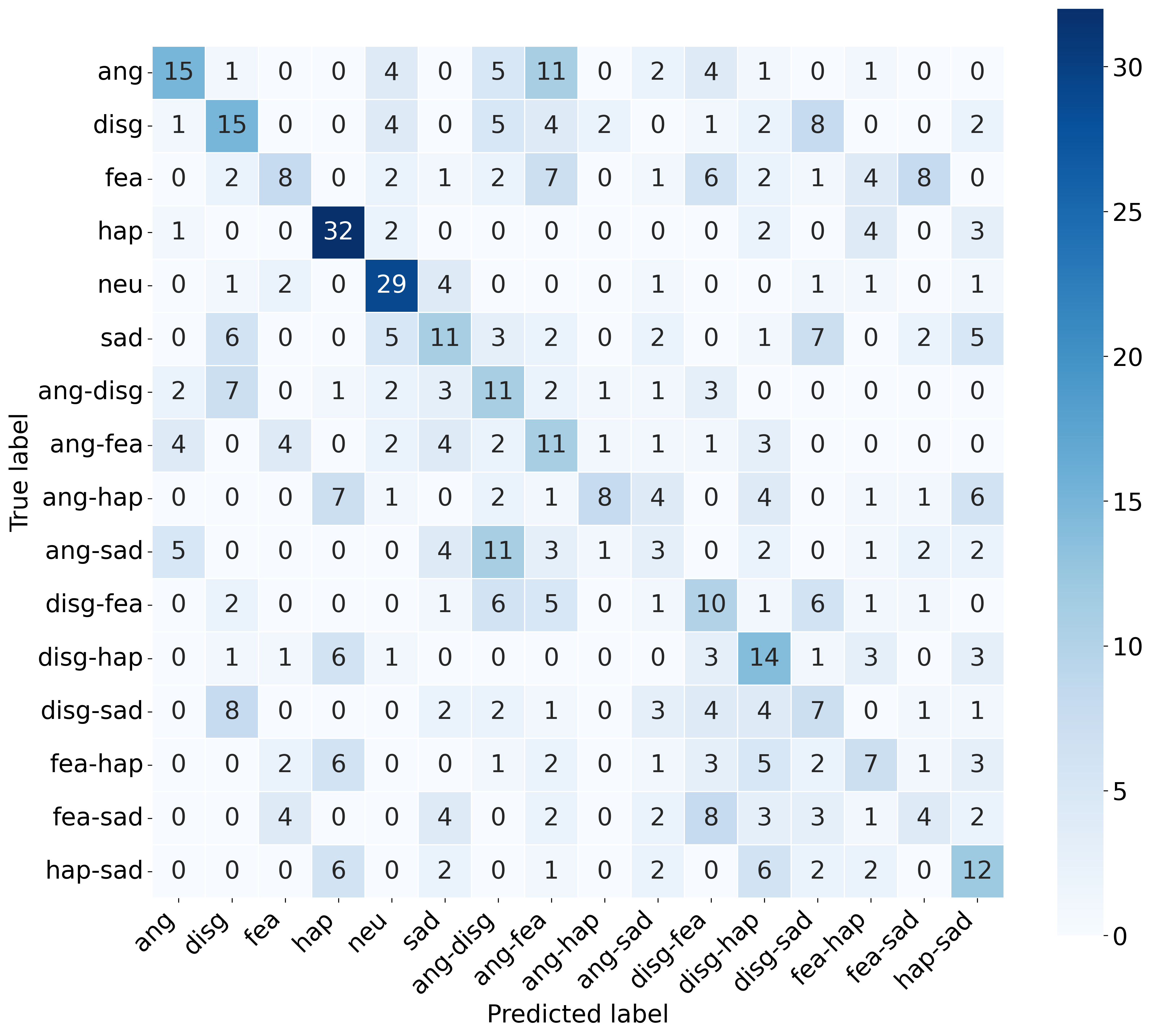}
    \caption{Confusion matrix for the test set using the best overall model (VideoMAEv2 + HuBERT, Aggregation).}
    \label{fig:confusion-matrix}
\end{figure}

To better understand prediction patterns on the test set, we visualized a confusion matrix based on the post-processed outputs of the best-performing model (VideoMAEv2 + HuBERT), see Figure~\ref{fig:confusion-matrix}. Each sample was assigned a single composite label corresponding to its ground-truth emotion configuration (e.g., \textit{happiness-sadness}, \textit{neutral}). This allows for a qualitative inspection of how well the model distinguishes between single and blended emotions.

\subsection{Feature Visualization}

\begin{figure}[h]
    \centering
    \begin{subfigure}{0.49\textwidth}
        \includegraphics[width=\linewidth]{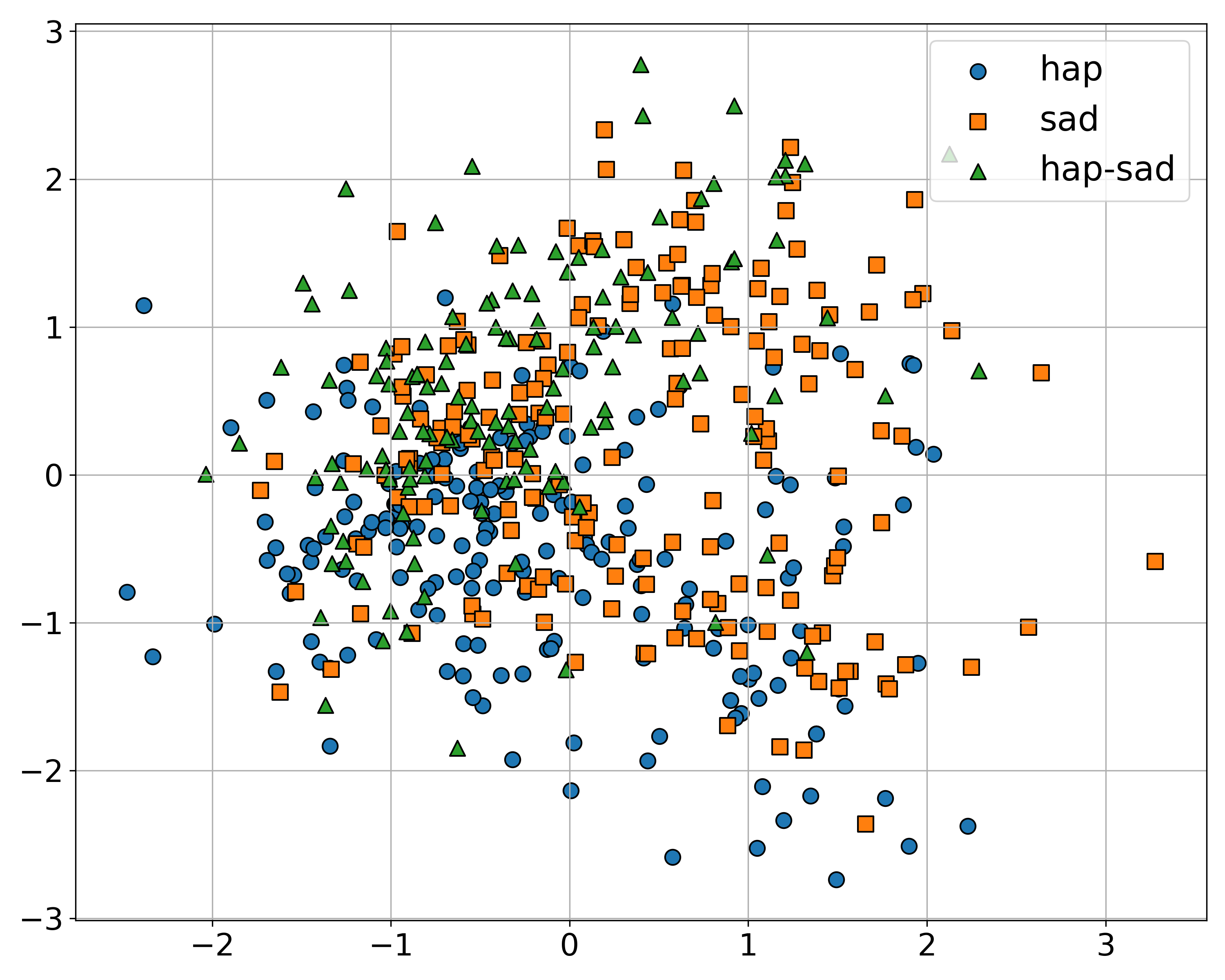}
        \caption{WavLM embeddings.}
        \label{fig:projector-imagebind}
    \end{subfigure}
    
    \vspace{0.3cm} %

    \begin{subfigure}{0.49\textwidth}
        \includegraphics[width=\linewidth]{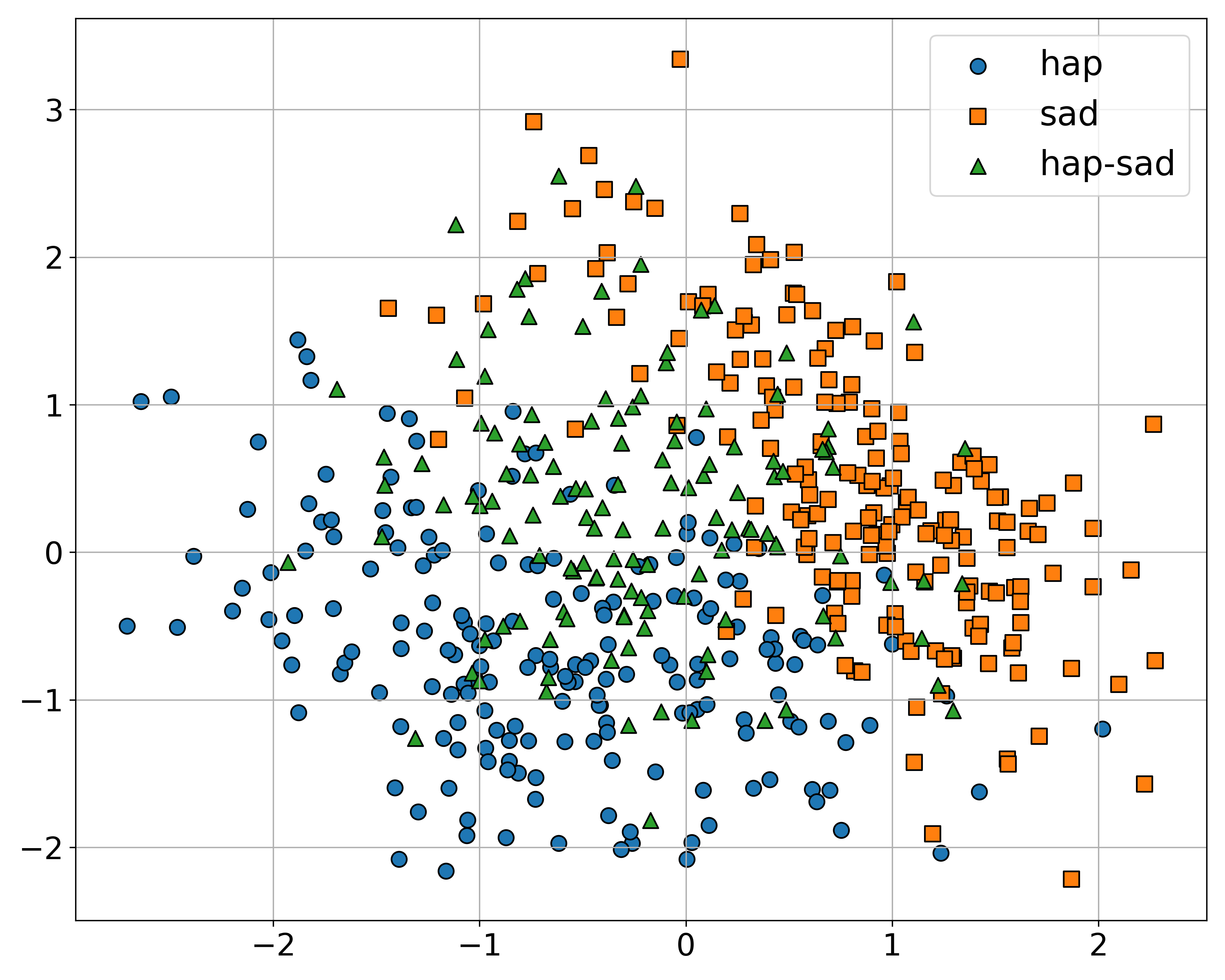}
        \caption{VideoMAEv2 embeddings.}
        \label{fig:projector-videomae}
    \end{subfigure}
    
    \caption{2D PCA projection of embeddings for the happy and sad emotions.}
    \label{fig:projector-visualization}
\end{figure}

To qualitatively explore the structure of the encoded representations, we applied PCA (Principal Component Analysis) to the aggregated embeddings obtained from the WavLM and VideoMAEv2 encoders. For visualization purposes only, we applied per-actor normalization that standardizes features within each individual actor. This reduces actor-specific biases and prevents clustering based on identity, which otherwise dominates the projection space. The resulting 2D plots are shown in Figure~\ref{fig:projector-visualization} for samples labeled with \textit{happy}, \textit{sad}, or their blends. The clusters show distinct patterns corresponding to single and blended emotion portrayals.

\section{Discussion}

In summary, our results show that multimodal approaches achieved the highest accuracies on the test set. The combination of VideoMAEv2 \cite{videomaev2} and HuBERT \cite{hsu2021hubert} reached an \accpresence{} of 0.332, while HiCMAE \cite{sun2024hicmae} achieved an \accsalience{} of 0.180. Although these results leave room for improvement, they are clearly above the trivial baselines for these challenging multi-class tasks (0.074 for \accpresence{} and 0.033 for \accsalience{}). To place these results in context, we next consider findings from recent work on the MAFW \cite{liu_mafw_2022} dataset. AVF-MAE++ \cite{wu2025avf} reported an unweighted average recall (UAR) of 17.25\% and a weighted average recall (WAR) of 43.83\%, with HiCMAE \cite{sun2024hicmae} achieving 13.29\% UAR and 37.36\% WAR on the same 43-class compound-emotion task. Given the strong class imbalance in MAFW, UAR provides a more informative measure of performance, particularly for blended emotions. While the two setups are not directly comparable, the best \accpresence{} result of 0.332 in our setting is roughly similar in scale to the reported UAR values, suggesting comparable task difficulty.

Turning to our own experiments, several factors help explain the observed performance characteristics. The observed performance drop from validation to test is primarily attributable to post-processing sensitivity rather than to training instability. While all models reached convergence, the thresholding procedure used to convert soft predictions into discrete presence indicators and salience levels introduces variability across datasets and training epochs. 

The models are trained to minimize Kullback–Leibler divergence, which aligns predictions with soft target distributions. For evaluation, however, these outputs must be discretized using fixed thresholds for presence and salience. In the validation evaluation, the presence and salience thresholds ($\alpha$ and $\beta$) are tuned to maximize performance on the held-out validation set, introducing some look-ahead bias. On the test set, however, these thresholds remain fixed. This makes the evaluation highly sensitive: even small shifts in the predicted distributions can cause substantial drops in accuracy. This is evident in the test performance of the VideoMAEv2 Aggregation model. The \accpresence{} is high while the \accsalience{} is close to trivial performance. Likely, a high presence threshold ($\alpha$) allowed the model to minimize the impact of false positives, yielding high \accpresence{}, while also predicting fewer blended emotions, which in turn yielded a lower \accsalience{}.

These issues illustrate the broader challenges of blended emotion recognition when it is modeled as a multi-label soft classification task. Alternative formulations may offer more stable solutions. For example, regression-based approaches can predict salience proportions directly, ranking-based formulations can capture relative prominence without requiring hard thresholds, and multi-task setups can jointly optimize presence and salience. Recent work has also explored alternative training objectives beyond standard cross-entropy, such as the bi-center loss~\cite{dong2024bi}, which anchors representations to the base emotions. However, despite progress, most state-of-the-art multimodal models~\cite{wu2025avf, zhang2023transformer, sun2024hicmae, li2024pth} have not been tailored specifically to blended emotion recognition, highlighting the need for further methodological advances.
\subsection{Limitations}
While \cn{} provides the largest and most balanced dataset of blended emotion expressions with salience annotations to date, several limitations should be acknowledged and addressed in future research. First, the expressions were acted in controlled laboratory conditions, which may limit ecological validity compared to spontaneous interactions in the wild \cite{juslinSpontaneousVocalExpressions2021}. Previous research has reported subtle yet systematic differences between acted and spontaneous expressions, e.g., \cite{juslinMirrorOurSoul2018}, \cite{nambaSpontaneousFacialExpressions2017}. Second, the emotion set is restricted to five basic categories and their pairwise blends. Although these particular blends are frequently reported in research \cite{Oatley1994}, real-world emotion blends may in addition involve more complex or culturally specific combinations. Third, salience was expressed only at three levels (50/50, 70/30, 30/70). This provides an initial demonstration of how relative prominence can vary in blended emotion expressions, but also reduces what is likely a continuous variable into a limited set of categories. Fourth, the dataset has limited representation of ethnic and cultural diversity, as it was developed within a European context. Finally, we primarily employed existing encoders combined via standard fusion techniques to provide initial performance baselines. Future studies could employ tailored models based on more sophisticated fusion mechanisms to increase classification accuracy.

\section{Conclusion}
We introduced \cn{}, the first publicly available multimodal dataset that systematically captures blended emotion expressions with relative salience annotations. 
We established reference benchmarks for two core tasks: (1) predicting emotion presence (\accpresence{}) and (2) relative salience (\accsalience{}). 
We implemented and evaluated a classification pipeline using state-of-the-art video, audio, and multimodal encoders. Unimodal encoders were evaluated separately and with early fusion techniques. 

Our results show that combining modalities consistently gives higher accuracy, highlighting the complementary nature of facial and vocal cues in conveying emotional blends. 
Recognition of blended emotions from multimodal expressions remains an emerging field with broad potential applications, and we hope that the \cn{} dataset will stimulate further research into how blended emotions are conveyed through the face and voice, and how they can be effectively classified.

{\small
\bibliographystyle{ieee}
\bibliography{egbib} %
}

\end{document}